\pdfoutput=1
\documentclass[journal,twoside,web]{ieeecolor}
\usepackage{generic}
\usepackage{cite}
\usepackage{amsmath,amssymb,amsfonts,bm}
\usepackage{algorithmic}
\usepackage{graphicx}
\usepackage{algorithm,algorithmic}
\usepackage{hyperref}
\hypersetup{hidelinks=true}
\usepackage{textcomp}
\usepackage{booktabs}       
\usepackage{multirow}
\usepackage[compatibility=false]{caption}

\usepackage{array}         
\usepackage{amsmath}       
\usepackage{amsfonts}      
\usepackage{amssymb}       
\usepackage{bbm} 
\usepackage{colortbl}
\usepackage{graphicx}      
\usepackage{float}        
\usepackage{colortbl}      
\usepackage{caption} 
\usepackage{booktabs,array,makecell}

\usepackage{eso-pic}
\AddToShipoutPictureBG*{%
  \AtPageUpperLeft{%
    \raisebox{-0.5cm}{\makebox[\paperwidth]{\scriptsize
      Accepted for publication in IEEE Journal of Biomedical and Health Informatics. DOI: \href{https://doi.org/10.1109/JBHI.2026.3693747}{10.1109/JBHI.2026.3693747}}}}%
  \AtPageLowerLeft{%
    \raisebox{0.5cm}{\makebox[\paperwidth]{\parbox[b]{0.85\paperwidth}{\scriptsize\centering
      \copyright~2026 IEEE. Personal use of this material is permitted. Permission from IEEE must be obtained for all other uses, in any current or future media, including reprinting/republishing this material for advertising or promotional purposes, creating new collective works, for resale or redistribution to servers or lists, or reuse of any copyrighted component of this work in other works.}}}}}
\def\BibTeX{{\rm B\kern-.05em{\sc i\kern-.025em b}\kern-.08em
    T\kern-.1667em\lower.7ex\hbox{E}\kern-.125emX}}
\newcommand{\gy}[1]{\textcolor{black}{#1}}

\begin{document}
\title{GTAFL: Addressing Test-Agnostic Long-Tailed Federated Learning in Medical Image}
\author{Guangyu Chen, Jingyun Zeng, Hui Jiang
\thanks{This work is supported by the fund which aims to improve scientific research capability of key construction disciplines in Guangdong province “Light-weight federal learning paradigm and its application” (No:2022ZDJS058); (Corresponding author: Hui Jiang)}
\thanks{Hui Jiang is with the School of Computer Science and Engineering, Huizhou University, Huizhou, China (e-mail: jh@hzu.edu.cn). Guangyu Chen is with the College of Engineering, Northeastern University, Boston, MA 02115, USA (e-mail: chen.guangyu1@northeastern.edu). Jingyun Zeng is with the College of  Computer Science, Beijing Institute of Technology, Zhuhai, China (e-mail:3120256743@bit.edu.cn)}
}

\maketitle

\begin{abstract}
In healthcare, protecting patient privacy is crucial due to the sensitivity of medical data and its extensive accessibility. Federated Learning (FL) offers a decentralized and privacy-preserving training paradigm, making it an ideal solution for healthcare applications. A critical challenge in healthcare is that real-world medical data often exhibits long-tailed distributions in local and global views. Existing methods addressing long-tailed FL problem typically assume that the model will be evaluated on uniform test data distribution. However, Practical test data in healthcare systems is often agnostic and unpredictable, leading to potential model failures in real-world scenarios. In this paper, we introduce a novel task termed Test-Agnostic Long-Tailed Federated Learning and propose GTAFL, a comprehensive framework to address this challenge. During the training stage, GTAFL employs adaptive re-sampling, expert classifier retraining, and self-supervised learning to correct biased classifiers and distorted feature spaces caused by long-tailed training distributions. During the inference stage, an ensemble mechanism combines retrained expert classifiers to handle test data with unknown distributions. Extensive experiments on CIFAR10 and two medical datasets manifest that our framework outperforms other state-of-the art methods. The code is available at {\color{blue}\url{https://github.com/cgy-lavine/GTAFL}}
\end{abstract}


\section{Introduction}
Recently, the rapid proliferation of Internet of Things (IoT) technologies has significantly expanded the application of deep learning techniques for processing heterogeneous data from diverse sources. The Internet of Medical Things (IoMT), a specialized subset of IoT infrastructure interconnected through computer networks \cite{IoT}, enables comprehensive data collection for patient monitoring and treatment optimization\cite{lian2022deep}. However, data privacy protection has emerged as a critical concern in IoMT applications. Federated Learning (FL) addresses this challenge by allowing multiple nodes to collaboratively train powerful models without exposing raw data~\cite{mcmahan2017communication,10589503,jiang2024lancelot,lian2022blockchain}, offering a privacy-preserving training strategy with considerable potential in medical analysis \cite{jiang2022harmofl,dou2021federated,gadekallu2023guest,jiang2022dynamic}.   

\begin{figure}[!t]
    \centering
    \includegraphics[width=\columnwidth]{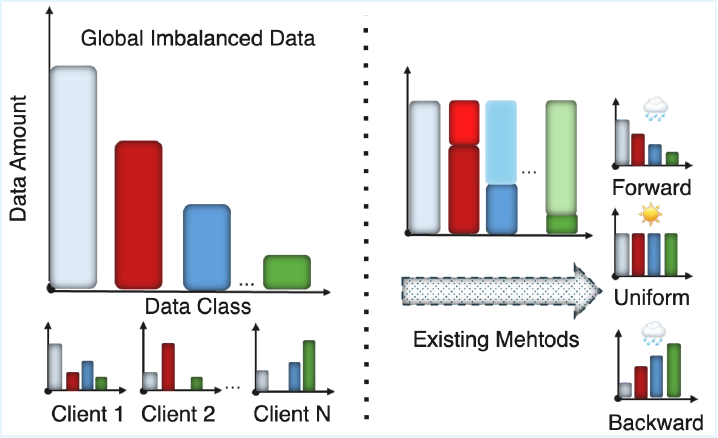}
    \caption{The illustration of two practical challenges: class imbalance and test agnostics. Existing method is only focused on uniform tests in federated settings.}
    \label{fig:existing}
\end{figure}

\gy {However, real-world federated learning (FL) faces significant challenges, notably class imbalance, which is common in medical datasets due to the rarity of certain conditions\cite{ye2026fedgsca}}. Such imbalance often biases models toward head classes, leading to poor performance on tail classes~\cite{Cai2021ACEAC, Zhang2021DeepLL}. To mitigate this, several studies \cite{Hou2025CAPTCP,li2025fedwcmunleashingpotentialmomentumbased,pmlr-v260-lu25a} have investigated long-tailed recognition in federated settings. \gy{However, existing methods assume that the test data is uniformly distributed. As illustrated in Fig.~\ref{fig:existing}, this assumption rarely holds in practice as the test distribution is often unknown. For example, a model trained primarily on  healthy individuals catastrophically degrades during an epidemic when the proportion of patient data surges. We refer to this practical yet underexplored task as \textit{Test-Agnostic Long-Tailed Federated Learning}. To the best of our knowledge, this task has not been previously studied in the federated learning literature. \footnote{We provide a problem formulation of test-agnostic long-tailed federated learning in \S \ref{sec:problem_formulation}.}}

To address this task, two primary challenges emerge. First, class imbalance manifests at both global and local levels. Globally, the aggregated client datasets typically exhibit long-tailed distributions where head classes substantially outnumber tail classes, despite the critical importance of rare classes in safety-sensitive domains such as medical diagnosis \cite{10869336}. Consequently, FL models demonstrate degraded performance on infrequent classes due to insufficient representation during training. Locally, imbalance occurs at the individual client level, where certain classes dominate specific clients while remaining scarce or entirely absent on others, thereby exacerbating inter-client heterogeneity (non-IID data). This distributional inconsistency results in biased aggregation, as clients develop divergent feature representations and decision boundaries for identical classes, ultimately compromising global model performance \cite{wang2021addressing,huang2024federated}. Although recent FL approaches, including re-sampling strategies \cite{shuai2022balancefl,10525198,10825694}, classifier re-training \cite{shang2022federated}, and feature-space adjustments \cite{Xiao2024FedLoGeJL}, attempt to mitigate these imbalances, they predominantly assume uniform test distributions, an assumption that rarely holds in practice.

The second challenge lies in generalization to unknown test distributions. In centralized training, methods can leverage prior knowledge for post-hoc adjustments \cite{Hong2020DisentanglingLD} or train multiple expert models \cite{Zhang2021SelfSupervisedAO} to accommodate various test distributions. However, FL settings impose fundamental constraints that render such approaches impractical. Resource limitations on edge devices preclude training multiple models locally\cite{10825769,jiang2024artfl}, while the distributed nature of FL complicates adaptation to unpredictable distribution shifts. For instance, while SADE \cite{Zhang2021SelfSupervisedAO} successfully trains multiple expert models with different loss functions for ensemble prediction, this approach fails in FL contexts where class absence at individual clients results in poorly trained experts, thereby compromising aggregation and  ensemble effectiveness. Consequently, practical FL solutions must achieve robustness to shifting test distributions while operating within inherent constraints imposed by client resource limitations and heterogeneous training distributions.

Therefore, we propose GTAFL, a framework for relieving test-agnostic long-tailed federated learning. Rather than requiring resource-constrained clients to train multiple potentially biased expert models, GTAFL leverages global information, specifically the aggregated training distribution, to align local and global training distributions through a global information-guided sampling strategy. Additionally, this global information guides the generation of synthetic data (termed global federated features), which enables the server to train multiple unbiased expert classifiers. This design draws inspiration from two-stage training paradigms \cite{shang2022federated,Kang2019DecouplingRA}, wherein a fixed feature extractor and balanced data are used to retrain biased classifiers. \gy{To better capture intrinsic visual features under unknown test distributions, we employ server-side SimCLR~\cite{Chen2020ASF} on unlabeled test 
set, without exposing any personally identifiable metadata~\cite{jiang2022pgada}. Since SimCLR is purely self-supervised and never accesses label information, this does not constitute data leakage, consistent with established FL works that leverage unlabeled server-side data for model refinement~\cite{fedds,fedmatch}.} These sampling strategies and strengthened representations, combined with classifier retraining and expert ensemble during the inference stage, enable GTAFL to maintain robustness under unknown and shifting test distributions. Extensive experiments on three datasets demonstrate that GTAFL outperforms state-of-the-art federated learning methods across various data distributions. The main contributions of this paper are summarized as follows:

\begin{itemize}
\item We introduce and analyze a new federated learning setting, i.e., test-agnostic long-tailed federated learning, in which models must contend simultaneously with long-tailed training data and unknown, potentially shifting test-time distributions.
\item We present GTAFL, a novel framework designed to address this challenge through four key components: adaptive re-sampling, expert classifier retraining, self-supervised learning, and ensemble learning.
\item We investigate synthetic global federated features and find that mild distributional imbalance does not significantly impair classifier retraining. Experiments show GTAFL achieves 8.5\% average improvements in backward test distributions while maintaining competitive performance in forward and uniform scenarios.
\end{itemize}
\section{Preliminaries and Problem Formulation}

\begin{table}[t]\small
    \centering
    \label{tab:notation}
    \caption{Notation Table}
    \begin{tabular}{ll}
    \toprule
    \textbf{Symbol} & \textbf{Description} \\
    \midrule
    $K$ & Number of federated clients \\
    $C$ & Number of classes \\
    $\mathbf{w} = \{\mathbf{u}, \mathbf{v}\}$ & Global model parameters \\
    $\mathbf{u}^t$ & Feature extractor parameters at round $t$ \\
    $\mathbf{v}^t$ & Original classifier parameters at round $t$ \\
    $\hat{\mathbf{v}}^t$ & Expert classifiers parameters at round $t$ \\
    $\hat{\mathbf{v}}_{f}, \hat{\mathbf{v}}_{u}, \hat{\mathbf{v}}_{b}$ & Forward, uniform, backward experts \\
    $f_{\mathbf{u}}$ & Feature extractor function \\
    $h_{\mathbf{v}}$ & Classifier function \\
    $m_c^{(i)}$ & Number of class $c$ samples on client $i$ \\
    $\boldsymbol{d}_{\text{local}}^{(i)}$ & Local class-count vector for client $i$ \\
    $\boldsymbol{d}_{\text{global}}$ & Global class-count vector \\
    $\mathbf{IF}$ & Imbalance factor \\
    $\mathcal{A}^t$ & Selected clients subset at round $t$ \\
    $w_c^i$ & Sampling weight for class $c$ on client $i$ \\
    $\alpha_i$ & Adaptive mixing parameter for client $i$ \\
    $\mathbf{g}_{c,\hat{\mathbf{v}}_{k}^t}^{j}$ & Class $c$ gradient from client $j$ for expert $k$ \\
    $\mathbf{g}_{c,\hat{\mathbf{v}}_{k}^t}^{\mathrm{agg}}$ & Aggregated class $c$ gradient for expert $k$ \\
    $\mathbf{g}_{c,\hat{\mathbf{v}}_{k}^t}^{\mathrm{gff}}$ & Class $c$ gradient from global federated features \\
    $\mathbf{F}_{c,\hat{\mathbf{v}}^t_{k}}^{t}$ & Global federated features for class $c$, expert $k$ \\
    $\tilde{m}_c$ & Global federated features count for class $c$ \\
    $w_k$ & Confidence weight for expert $k$ \\
    \bottomrule
    \end{tabular}

\end{table}
\subsection{Federated Learning}
Federated learning (FL) aims to obtain a global model without sharing local data. In a typical FL setting, suppose that we have $K$ clients, each indexed by $k \in \{1, 2, \ldots, K\}$. Each client $k$ holds a local dataset $\mathcal{D}^k = \{(x_i, y_i)\}_{i=1}^{N^k}$, where $N^k$ denotes the number of samples on the $k$-th client. Client datasets are often heterogeneous and non-independent and identically distributed (non-IID). The goal of FL is to collaboratively learn a global model $\phi_{\mathbf{w}}$ over the union of all local datasets $\mathcal{D} = \bigcup_{k=1}^K \mathcal{D}^k$ with the coordination of a central server, while preserving data privacy by avoiding direct data transmission. The global model $\phi_{\mathbf{w}}$ is parameterized by learnable weights $\mathbf{w} = \{\mathbf{u}, \mathbf{v}\}$ and typically consists of two components: (1) a feature extractor $f_{\mathbf{u}}: \mathcal{X} \to \mathcal{Z} \subseteq \mathbb{R}^d$ with parameters $\mathbf{u}$, which maps each input $\mathbf{x} \in \mathcal{X}$ to a $d$-dimensional feature vector $\mathbf{z} = f_{\mathbf{u}}(\mathbf{x}) \in \mathcal{Z}$; (2) a classifier $h_{\mathbf{v}}: \mathcal{Z} \to \mathbb{R}^{|\mathcal{I}|}$ with parameters $\mathbf{v}$, which maps a feature vector $\mathbf{z} \in \mathcal{Z}$ to a vector of logits. It is typically implemented as a fully connected layer that outputs class confidence scores, where $\mathcal{I} = {1,2,\ldots,C}$ denotes the set of $C$ classes.

\subsection{Execution Progress of Federated Learning}
In FL, the training process consists of multiple communication rounds, where each round $t$ involves the following two steps. The first step is the local model training. At the beginning of round $t$, the central server broadcasts the current global model parameters $\mathbf{w}^t$ to a selected subset of clients $\mathcal{A}^t \subseteq \{1, 2, \ldots, K\}$. Suppose that the parameters of the $k$-th client's local model are denoted as $\mathbf{w}_k$. Each participating client $k \in \mathcal{A}^t$ receives the global model and performs local training on their private dataset $\mathcal{D}^k$ using gradient descent. The local model update is given by:
\begin{equation}
    \mathbf{w}_k^{t+1} \leftarrow \mathbf{w}_k^t - \eta \nabla_{\mathbf{w}} \ell(\mathbf{w}_k^t; \mathcal{D}^k) 
\end{equation}
where $\ell$ is cross-entropy loss.

The second step is the model aggregation. After local training, the participating clients upload their updated model parameters $\mathbf{w}_k^{t+1}$ to the server. The server then performs a weighted averaging operation to update the global model for the next communication round:
\begin{align}
\mathbf{w}^{t+1} = \sum_{k \in \mathcal{A}^t} \frac{|\mathcal{D}^k|}{\sum_{j \in \mathcal{A}^t} |\mathcal{D}^j|} \mathbf{w}_k^{t+1} 
\end{align}
Note that we use FedAvg~\cite{mcmahan2017communication} as our aggregation method, which serves as the foundation for most federated learning approaches.

\subsection{Local and Global Data Distributions}

Let $m_c^{(i)}$ denote the number of samples from class $c \in \{1,\dots,C\}$ on client $i \in \{1,\dots,K\}$. 
The local class-count vector for client $i$ is formally defined as
\begin{equation}
    \boldsymbol{d}_{\text{local}}^{(i)} = \bigl(m_1^{(i)}, \dots, m_C^{(i)}\bigr) \in \mathbb{R}_+^{C}
\end{equation}

The global class-count vector, which aggregates the class distributions across all $K$ participating clients, is given by
\begin{equation}
\boldsymbol{d}_{\text{global}} = \left( \sum_{i=1}^{K} m_1^{(i)}, \dots, \sum_{i=1}^{K} m_C^{(i)} \right) \in \mathbb{R}_+^{C}
\end{equation}
To quantify the degree of class imbalance inherent in the global distribution, we introduce the imbalance factor (IF), also defined as the imbalanced ratio between the maximum and minimum global class counts:
\begin{equation}
   \mathbf{IF} \;=\; 
\frac{\max_{c \in \{1,\dots,C\}} \sum_{i=1}^{K} m_c^{(i)}}
     {\min_{c \in \{1,\dots,C\}} \sum_{i=1}^{K} m_c^{(i)}}
\end{equation}

\subsection{Problem Formulation}
\label{sec:problem_formulation}
We provide a problem formulation of test-agnostic long-tailed federated learning, where there exists distribution discrepancy between training and testing data. Specifically, the global training set $\mathcal{D}_{\mathrm{train}}$ follows a long-tailed distribution characterized by the imbalance factor $\mathbf{IF}$, while the test set $\mathcal{D}_{\mathrm{test}}$ has an unknown class distribution that may differ significantly from the training distribution.

Let $\mathcal{D}_{\mathrm{test}}$ with class counts $(m_1, m_2, \ldots, m_C)$ where each $m_c$ is unknown and potentially different from the training class frequencies. The test distribution can follow forward, uniform, or backward patterns relative to the training long-tail.

Our objective is to learn a robust global model that performs well across different potential test distributions:
\begin{equation}
   \min_{\mathbf{w}} \mathbb{E}_{\mathcal{D}_{\mathrm{test}}} \left[ \frac{1}{|\mathcal{D}_{\mathrm{test}}|} \sum_{(x_i,y_i) \in \mathcal{D}_{\mathrm{test}}} \ell(\phi_{\mathbf{w}}(x_i), y_i) \right] 
\end{equation}

where the expectation is over the unknown test distribution and $\ell$ is the cross-entropy loss.

\section{Motivation and Overview}

\subsection{Motivation Study}

\begin{figure}[!t]
    \centering
    \includegraphics[width=\columnwidth]{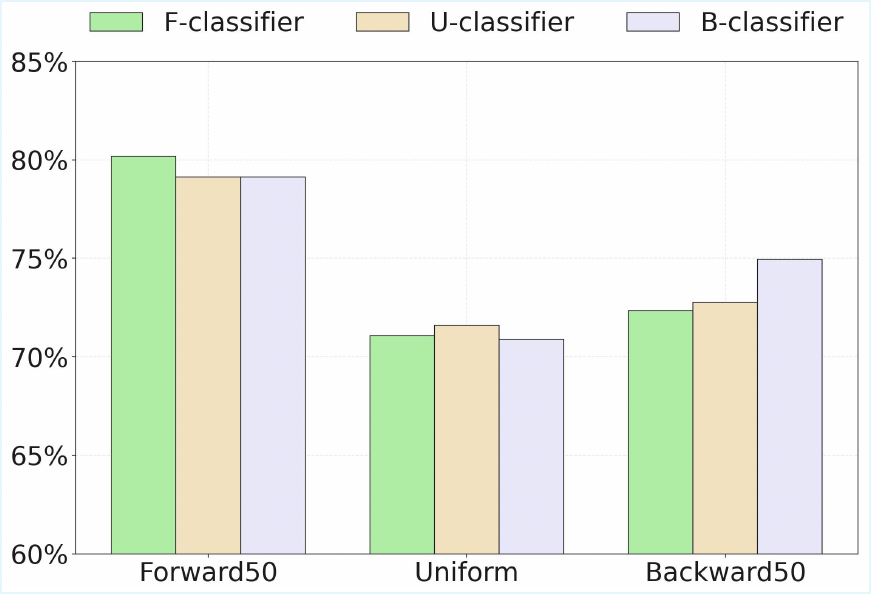}
    \caption{Expert classifier specialization demonstrates that each classifier achieves optimal performance on its corresponding test distribution while maintaining competitive accuracy across other test distribution.}
    \label{fig:Expert_classifier}
\end{figure}

In centralized training, long-tailed data distributions and distribution shifts between training and testing data typically degrade model performance\cite{wei2024learning,Yang2024HarnessingHL,ye2023licam,jiang2025large}. In federated learning (FL) with imbalanced data, such shifts present even greater challenges due to data heterogeneity across clients. Previous work\cite{Xiao2024FedLoGeJL,Zhang2022FederatedLW} demonstrates that in long-tailed FL scenarios, fine-tuning only the classifier can effectively mitigate long-tailed bias without complete model retraining.
The key challenge lies in training multiple expert classifiers to handle various tests with unknown distribution. Due to privacy constraints inherent in federated learning, servers cannot directly access training data. Consequently, auxiliary datasets or synthetic data are commonly used for expert model training. CReFF\cite{shang2022federated} exemplifies this approach by leveraging federated features, balanced synthetic data trained from client-transmitted gradients, to retrain biased classifiers. Since federated features are  balanced, a natural extension involves modifying their distribution to enable optimal classifier performance on categories with synthetic majority representation.
Our experimental analysis on CIFAR10-LT, illustrated in Figure~\ref{fig:Expert_classifier}, reveals that retraining classifiers on federated features with different distribution yields varying performance in different tests. Specifically, classifiers retrained on distribution-specific federated features demonstrate superior performance on their corresponding test distributions. For instance, the backward-trained classifier achieves exceptional performance on backward test distributions while maintaining competitive accuracy across other distribution types.

\begin{figure}[!t]
    \centering
    \includegraphics[width=\columnwidth]{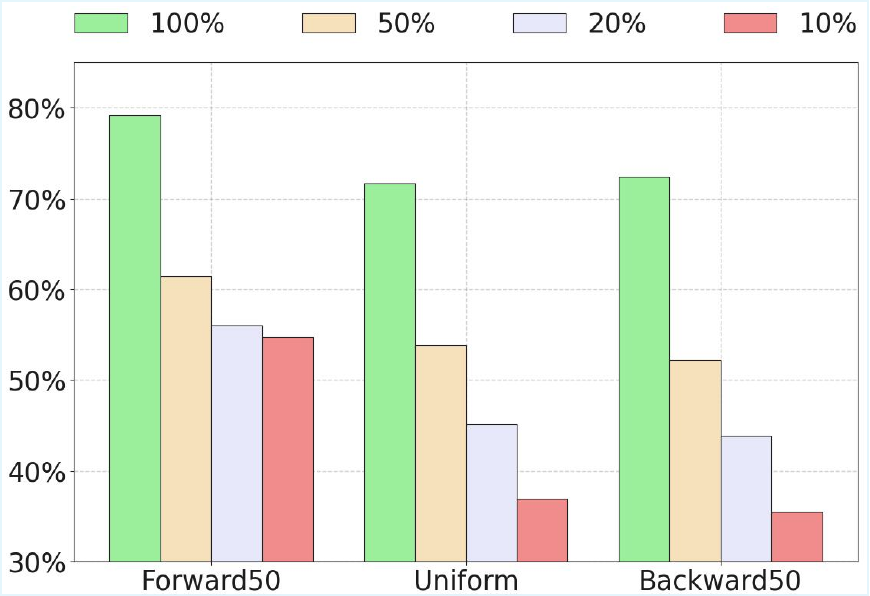}
    \caption{CReFF performance degradation with reduced training data size across different test distributions. Performance drops significantly as training data decreases from 100\% to 10\%, with Backward50 test showing the most severe degradation}
    \label{fig:discrepancy}
\end{figure}

The train-test distribution discrepancy becomes more pronounced when data is limited, as demonstrated in Fig.~\ref{fig:discrepancy}. In medical imaging, particularly during the pandemic, substantial volumes of unlabeled imaging data have become available. To effectively leverage this data while addressing the high annotation costs and stringent privacy requirements inherent to medical applications, we employ a self-supervised contrastive learning approach\cite{Li_2021_CVPR,jiang2022pgada}.
Specifically, we implement the SimCLR framework\cite{Chen2020ASF} on the server side to enable efficient feature learning from unlabeled medical images. This approach is particularly well-suited for federated medical scenarios as it exclusively utilizes image data without accessing any patient information or other sensitive metadata. The self-supervised learning process focuses solely on extracting intrinsic visual features, thereby maintaining the privacy-preserving principles fundamental to federated learning while enhancing the model's representational capabilities on medical imaging tasks.

\subsection{Overview of GTAFL }
\begin{figure*}[h]
\centering
\includegraphics[width=18cm]{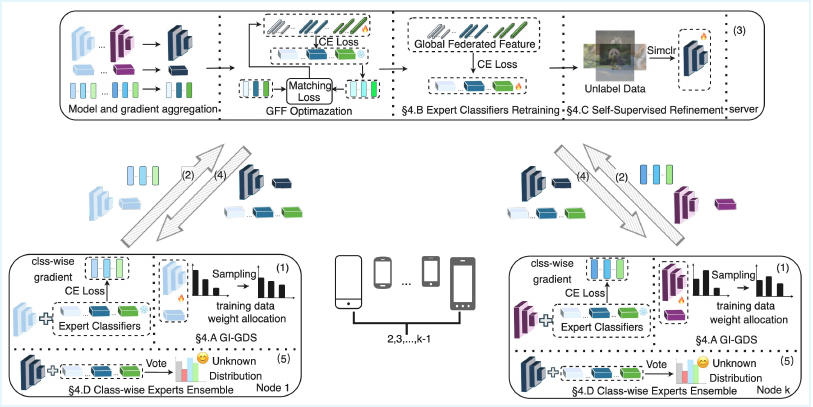}
\caption{GTAFL framework workflow: (1) Clients receive global model and expert parameters, then perform local training with GI-GDS sampling and gradient computation. (2) Clients transmit class-wise gradients and model parameters to the server. (3) Server performs aggregation, expert retraining using optimized global federated features, and self-supervised refinement with unlabeled data. (4) The server sends the updated global model and expert classifiers parameters to clients. (5) Expert ensemble inference for test-agnostic prediction.}
\label{fig:framework}
\end{figure*}

The primary goal of GTAFL is to train specialized expert classifiers and effectively combine them during the inference stage to handle varying test distributions. \gy{As illustrated in Fig. \ref{fig:framework}, the proposed GTAFL framework consists of four components: data sampling, expert classifiers retraining, self-supervised refinement, and class-wise experts ensemble. Each communication round consists of the following steps: (1) Clients  train the received global model based on the GI-GDS sampling strategy while computing class-wise gradients with respect to the expert classifiers from server. (2) Clients transmit the updated local model and class-wise gradients to the server. (3) On the server side, the server aggregates feature extractor, original classifier and the class-wise gradients. The expert classifiers are retrained based on the optimized global federated features. The feature extractor is further improved based on self-supervised method.} (4) The server sends the aggregated model and retrained expert classifiers back to the clients, returning to stage (1). (5) During the inference stage, the expert classifiers with the shared feature extractor vote collaboratively to make predictions on test data.

\gy{As illustrated in Section~\ref{subsec:GI-GDS}, global information is leveraged for local data sampling to achieve a pseudo-balanced state during training. At this stage, only the feature extractor and the original classifier are updated locally. On the server side, the feature extractor and the original classifier are aggregated based on standard FedAvg approach~\cite{mcmahan2017communication}.} Expert classifier retraining uses synthetic data, specifically global federated features (GFF) in GTAFL. \gy{In order to optimize the GFF, matching loss\cite{zhao2020dataset} is used for minimizing the gradient distance between synthetic data and real data toward the same classifiers. Subsequently, expert classifiers are trained using these updated GFF to achieve specialization for specific data distributions. After retraining the expert classifiers, the feature extractor is further refined using unlabeled data through the SimCLR technique, enabling the model to learn robust and generalized feature representations. This process ensures strict data privacy, as no sensitive information such as patient names is exposed.}

\section{Methodology}

\subsection{Global Information–Guided Data Sampling (GI-GDS)}
\label{subsec:GI-GDS}

\gy{Sampling strategy that inverts the local class distribution can preserve minority-class signal on each client but often overcompensate, causing severe oversampling and noisy updates for rare classes~\cite{shuai2022balancefl}. Conversely, sampling solely from the global distribution can smooth extremes but may ignore client-specific information that is important for personalization.}

To balance these effects, we propose Global Information–Guided Data Sampling (GI-GDS), a hybrid sampling scheme that interpolates between local inverse-frequency and global inverse-frequency signals. For client $i$ and class $c$, let $m_c^i$ be the number of local samples in class $c$, and let $M_c=\sum_{j=1}^{N} m_c^j$ denote the total across all $N$ clients. The sampling weight is
\begin{equation}
w_c^i(\alpha) \;=\; \alpha \cdot \frac{1}{m_c^i + \varepsilon} \;+\; (1-\alpha)\cdot \frac{1}{M_c + \varepsilon}
\label{eq:gigds-weight}
\end{equation}
where $\varepsilon>0$ (e.g., $10^{-6}$) prevents division by zero. We use normalized weights $\tilde{w}_c^i = w_c^i / \sum_{c'} w_{c'}^i$ to form the per-client sampling distribution.

\gy{At the start of each communication round, each participating client uploads its local class-count vector $\boldsymbol{d}_{\text{local}}^{(i)}$
to the server. The server computes the global class-count vector $M_c$ by summation and broadcasts it back to all clients for use in the GI-GDS sampling weights. If the local class distribution $\boldsymbol{d}_{\text{local}}^i$ substantially differs from the global distribution $\boldsymbol{d}_{\text{global}}$, relying on global-only weights can erase informative local patterns.} Conversely, when the two distributions are similar, emphasizing the global signal helps avoid aggressive oversampling of very rare local classes.

\gy{Let $s_i = \cos\!\left(\boldsymbol{d}_{\text{local}}^i,\, \boldsymbol{d}_{\text{global}}\right)$ be the cosine similarity between the client’s  class-frequency vector and the global frequency vector. We set
\begin{equation}
\alpha_i \;=\;
\begin{cases}
1, & \text{if } s_i < \tau,\\
0, & \text{otherwise,}
\end{cases}
\qquad \text{with } \tau=0.9
\label{eq:gigds-alpha}
\end{equation}
Thus, when the local and global distributions diverge ($s_i<\tau$), we set $\alpha_i=1$ to preserve client-specific information; when they are similar, we set $\alpha_i=0$ to limit minority oversampling.}

\subsection{Expert Classifier Retraining}
\label{classifier retraining }
In long-tailed federated learning, a single global classifier often fails to generalize uniformly, especially under training-testing distribution mismatch. \gy{To mitigate this mismatch, we maintain multiple classifier experts to specialize for a target prior while sharing a common feature extractor.}

At communication round $t$, the server broadcasts global parameters $\{\mathbf{u}^{t},\mathbf{v}^{t}\}$ and expert classifiers parameters $\hat{\mathbf{v}}^{t}=\{\hat{\mathbf{v}}_{f_1}^{t},\ldots,\hat{\mathbf{v}}_{f_p}^{t},\hat{\mathbf{v}}_{u_1}^{t},\ldots,\hat{\mathbf{v}}_{u_q}^{t},\hat{\mathbf{v}}_{b_1}^{t},\ldots,\hat{\mathbf{v}}_{b_r}^{t}\}$ targeting forward, uniform, and backward long-tail priors, respectively. \gy{ Each client trains $\{\mathbf{u}^{t},\mathbf{v}^{t}\}$ locally following FedAvg. Furthermore, for class $c$, client $j$ samples $n_{c}^{j}$ instances and reports the average gradient of cross-entropy loss for every expert classifier. The gradient computation for each expert classifier parameters $\hat{\mathbf{v}}^t_{k} \in \hat{\mathbf{v}}^t$,  is formulated as}

\begin{equation}
\mathbf{g}_{c,\hat{\mathbf{v}}_{k}^t}^{j}
= \frac{1}{n_{c}^{j}} \sum_{i=1}^{n_{c}^{j}} \nabla_{\hat{\mathbf{v}}_{k}^t} \ell\left(h_{\hat{\mathbf{v}}_{k}^t}(f_{\mathbf{u}^{t}}(x_{c,i}^{j})), y_{c,i}^{j}\right)
\end{equation}

Upon receiving updates from clients $\mathcal{A}^{t}$, the server aggregates model parameters to obtain $\{\tilde{\mathbf{u}}^{\,t},\mathbf{v}^{t+1}\}$ and averages class-wise expert gradients :
\begin{equation}
\mathbf{g}_{c,\hat{\mathbf{v}}^t_{k}}^{\mathrm{agg}}
=
\frac{1}{|\mathcal{A}_{c}^{t}|}
\sum_{j \in \mathcal{A}_{c}^{t}}
\mathbf{g}_{c,\hat{\mathbf{v}}^t_{k}}^{j}
\end{equation}

\gy{To enable server-side expert retraining without client raw data, the server maintains global federated features $\mathbf{F}_{c,\hat{\mathbf{v}}^t_{k}}^{t} \in \mathbb{R}^{\tilde{m}_{c}\times d}$ for each class $c$ and expert $\hat{\mathbf{v}}^t_{k}$. These features are designed to generate gradients that approximate those from raw data. The gradients generated by GFF for $\hat{\mathbf{v}}^t_{k}$ and class $c$ are as following}:
\begin{equation}
\mathbf{g}_{c,\hat{\mathbf{v}}_{k}^t}^{\mathrm{gff}}
=
\frac{1}{\tilde{m}_{c}}
\sum_{i=1}^{\tilde{m}_{c}}
\nabla_{\hat{\mathbf{v}}_{k}^t}\,
\ell\left(
h_{\hat{\mathbf{v}}_{k}^t}(\mathbf{F}_{c,\hat{\mathbf{v}}_{k}^t,i}^{t}),
y_{c}
\right)
\end{equation}

The GFF's class-specific count is defined as $\tilde{m}_{c} = m + \lambda\,\tilde{\boldsymbol{d}}^c_{\mathrm{global}}$, where $m$ is a balanced base count and $\tilde{\boldsymbol{d}}_{\mathrm{global}}^{c}$ is the global class frequency. For the balanced expert, we set $\lambda=0$.

These features are refined by aligning their induced gradients to the aggregated client gradients. We optimize the GFF by minimizing the cosine matching loss:
\begin{equation}
D\left(
\mathbf{g}_{c,\hat{\mathbf{v}}_{k}^t}^{\mathrm{gff}},\,
\mathbf{g}_{c,\hat{\mathbf{v}}_{k}^t}^{\mathrm{agg}}
\right)
=
1
-
\frac{
\mathbf{g}_{c,\hat{\mathbf{v}}_{k}^t}^{\mathrm{gff}}
\cdot
\mathbf{g}_{c,\hat{\mathbf{v}}_{k}^t}^{\mathrm{agg}}
}{
\big\|\mathbf{g}_{c,\hat{\mathbf{v}}_{k}^t}^{\mathrm{gff}}\big\|\,
\big\|\mathbf{g}_{c,\hat{\mathbf{v}}_{k}^t}^{\mathrm{agg}}\big\|
}
\end{equation}

After refinement, we obtain $\mathbf{F}_{\hat{\mathbf{v}}_k^t}^{t}$ for all classes and retrain each expert parameters $\hat{\mathbf{v}}_{k}^{t}$ while freezing $\tilde{\mathbf{u}}^{\,t}$. The server updates the expert classifiers  as:
\begin{equation}
\hat{\mathbf{v}}_{k}^{t+1}
=
\hat{\mathbf{v}}_{k}^t
-
\eta\,
\nabla_{\hat{\mathbf{v}}_{k}^t}\,
\ell\left(
h_{\hat{\mathbf{v}}_{k}^t}(\mathbf{F}_{\hat{\mathbf{v}}^t_{k}}^{t}),
y
\right)
\end{equation}

\subsection{Self-Supervised Refinement}
\label{ssr}
\gy{After aggregating the feature extractor parameters $\tilde{\mathbf{u}}^{t}$ and completing expert retraining, we further refine the feature extractor using self-supervised refinement, i.e., contrastive learning, on unlabeled images available at test time. This step leverages medical images only, preserving the patients' privacy and federated learning protocol.}

\gy{At communication round $t$, the server draws $N$ unlabeled images and applies two stochastic augmentations to each. Each view forms one positive pair with its counterpart and $2N-2$ negatives with the remaining views in the batch.} The self-supervised refinement loss is defined as follows.
\begin{equation}
\mathcal{L}=\frac{1}{2N}\sum_{i=1}^{N}\big[\mathcal{L}_{2i-1,2i}+\mathcal{L}_{2i,2i-1}\big]
\end{equation}
$\mathcal{L}$ denotes the NT-Xent loss \cite{Chen2020ASF}

\begin{equation}
\mathcal{L}_{i,j}=-\log\frac{\exp(\mathrm{sim}(z_i,z_j)/\tau)}{\sum_{k=1, k \neq i}^{2N}\exp(\mathrm{sim}(z_i,z_k)/\tau)}
\end{equation}

where $\text{sim}({z_i}, \mathrm{z_j}) = {z_i \cdot z_j} /{\|z_i\|\|z_j\|}$ denote the dot product between $\ell_2$ normalized $z_i$ and $z_j$. $z_i$ is the projected feature of view $i$, and $\tau$ is the temperature.

We update the feature extractor by gradient descent,
\begin{equation}
\mathbf{u}^{t+1}=\tilde{\mathbf{u}}^{t}-\eta\nabla_{\tilde{\mathbf{u}}^{t}}\mathcal{L}
\label{simclr_update}
\end{equation}
using learning rate $\eta$ for this self-supervised phase, while keeping $\mathbf{v}^{t+1}$ and the experts $\hat{\mathbf{v}}^{t+1}$ fixed. The refined global model parameters  $\{\mathbf{u}^{t+1},\mathbf{v}^{t+1}\}$ and expert classifiers parameters  $\hat{\mathbf{v}}^{t+1}$ are then disseminated to clients for round $t+1$.

\subsection{Class-wise Experts Ensemble}
\label{ensemble}
\gy{After $t$ round training, GTAFL aggregates predictions from multiple experts through a hard voting mechanism to accommodate unknown test priors. Given an input $x$, features $z=f_{\mathbf{u}^{t}}(x)$ are extracted and evaluated by each expert classifier, which produces a discrete class prediction $\hat{y}_{k} = \arg\max_{c} h_{\hat{\mathbf{v}}_{k}^{t}}(z)$ by selecting the class with the highest logits score.}

We combine these discrete predictions through a weighted voting mechanism. For each candidate class $c$, we count the weighted votes from all experts who predict that class. The final prediction is determined by selecting the class that receives the maximum weighted vote count:
\begin{equation}
\hat{y} = \arg\max_{c} \sum_{k=1}^{p+q+r} w_{k} \cdot \mathbb{I}[\hat{y}_{k} = c]
\end{equation}
where $w_{k}\ge 0$ represents the confidence weight for the $k$-th expert, and $\mathbb{I}[\cdot]$ is the indicator function that equals 1 when the condition is true and 0 otherwise.

\gy{In the default implementation, we use uniform weighting ($w_{k}=1$ for all experts) with $p=q=r=1$, yielding one expert per distribution type. This democratic approach ensures equal contribution from each expert regardless of test distribution shifts, without requiring probability calibration or confidence estimation. The framework can be extended to support adaptive weighting schemes based on expert reliability or validation performance when such information is available.}

\section{Experiments Setup}

\begin{table*}[t]\small
\centering
\caption{Summary of datasets used in experiments.}
\begin{tabular}{c|c|c|c|c|c|c}
\toprule
\textbf{Dataset} & \textbf{Modality} & \textbf{Task} & 
\textbf{\makecell[c]{\# of TrainSet}} & \textbf{\# of Testset} & \textbf{\# of Classes} &\textbf{\makecell[c]{Imbalance \\ Ratio}} \\
\hline
CIFAR10-LT & RGB & Image Recognition & 50,000 & 10,000 & 10 & 50 \\
PathMNIST-LT & Colon Pathology & Med-Image Classification & 89,996 & 7,180 & 9 & 50 \\
OCTMNIST-LT & Retinal OCT & Med-Image Classification & 97,477 & 1,000 & 4 & 50 \\
\bottomrule
\end{tabular}

\label{datasets}
\end{table*}

\subsection{Datasets}

To evaluate GTAFL's effectiveness, we conduct experiments on three benchmark datasets: CIFAR10-LT, PathMNIST-LT, and OCTMNIST-LT. These datasets represent diverse domains including natural images and medical imaging applications. Table~\ref{datasets} summarizes the key characteristics of each dataset. We follow a standardized two-stage protocol for creating long-tailed federated datasets. First, we generate globally imbalanced datasets with an imbalance factor of 50 using exponential decay. Second, we partition datasets among clients using Dirichlet distribution with $\alpha = 0.5$ \cite{lin2020ensemble} to simulate non-IID federated environments. The following provides detailed specifications for each dataset.

\begin{itemize}
    \item \textbf{CIFAR10-LT}: Long-tailed variant of CIFAR-10 \cite{krizhevsky2009learning} containing 50,000 training images (32$\times$32 RGB) across 10 object categories, providing a standard computer vision benchmark for evaluating general image classification under federated long-tail conditions.
    
    \item \textbf{PathMNIST-LT}: Derived from the MedMNIST collection \cite{yang2023medmnist}, this colon pathology dataset comprises 89,996 training samples across 9 diagnostic classes with 7,180 test samples, representing realistic medical imaging scenarios with inherent class imbalance.
    
    \item \textbf{OCTMNIST-LT}: Derived from the MedMNIST collection \cite{yang2023medmnist}, OCTMNIST is a retinal optical coherence tomography dataset containing 97,477 training samples across 4 retinal condition categories with 1,000 test samples, providing validation for ophthalmological applications where rare conditions are clinically significant.
\end{itemize}

\subsection{Baselines}
In this section, we conduct extensive experiments to evaluate the performance of GTAFL, First, we evaluate the accuracy on  various test with different IF on three datasets. Next, we compare the class-wise performance on six baselines. Last but not least, we use an ablation study to find the effectiveness of resampling, classifier retraining, contrastive learning and ensemble strategies. In particular, the baseline includes traditional FL methods and other long-tailed oriented FL methods are compared with GTAFL.
\begin{itemize}
    \item \textbf{FedAvg}~\cite{mcmahan2017communication}: The canonical FL baseline that alternates local SGD on each client with server-side aggregation via data-size–weighted averaging of model parameters. Clients train on their private data for a fixed number of local steps, then upload updates to the server for synchronous aggregation.

    \item \textbf{FedProx}~\cite{li2020federated}: Extends FedAvg by adding a proximal (quadratic) regularization term to each client’s local objective, penalizing deviation from the current global model to mitigate client drift under non-IID data; it also accommodates variable amounts of local work and partial participation.

    \item \textbf{BalanceFL}~\cite{shuai2022balancefl}: Using resampling and model distillation techniques to solve label imbalance. Re-sampling strategy incorporates class-frequency–aware reweighting into the training and aggregation pipeline (e.g., balanced sampling/losses and balanced aggregation of client updates) to counter majority-class dominance.

    \item \textbf{CReFF}~\cite{shang2022federated}: employs a classifier retraining approach that generates class-wise synthetic features comparable to real features in a privacy-preserving manner. Server retrains the classifier using these synthetic features,  enabling effective classifier recalibration to mitigate long-tailed distribution bias. 
 .

    \item \textbf{FedGraB}~\cite{xiao2023fed}: Proposes gradient- and parameter-level rebalancing with regularization to dampen noisy or biased client updates; includes mechanisms that emphasize reliable contributions (e.g., by magnitude/consistency) while suppressing outliers caused by heterogeneity.

    \item \textbf{FedLoGe}~\cite{Xiao2024FedLoGeJL}: simultaneously optimizing both local and global models. It uses a shared backbone for global trends while employing individualized classifiers for client-specific features, achieving superior performance through representation learning and classifier alignment.
   
\end{itemize}

\subsection{Implementation Details} We use ResNet-8 as the backbone for all experiments. All methods are implemented in PyTorch and evaluated on a single NVIDIA GeForce RTX 4090 or two VGPU-48G GPUs. Federated learning is configured with 20 clients and 40\% client participation per round, running for 200 rounds on CIFAR10-LT and PathMNIST-LT, and 100 rounds on OCTMNIST-LT. For a fair comparison, all compared FL methods use the same model and optimizer. Each selected client trains locally for 5 epochs with a batch size of 32 for CIFAR10-LT, 1024 for PathMNIST-LT, and 64 for OCTMNIST-LT. and the server aggregates updates using {FedAvg}. On the server side, we initialize the number of global federated features per class to 100. We use SGD with learning rate 0.1 for all optimization. Unless otherwise noted, the global federated feature coefficient is set to $\lambda=0.01$ (CIFAR10-LT), $\lambda=0.001$ (PathMNIST-LT), and $\lambda=0.0001$ (OCTMNIST-LT). For SimCLR pretraining, the learning rates are $1\times10^{-3}$ (CIFAR10-LT), $1\times10^{-4}$ (PathMNIST-LT), and $1\times10^{-5}$ (OCTMNIST-LT), with 10, 2, and 2 epochs, respectively.

\section{Experiments}

\begin{table*}[t]\small
    \centering
    \caption{Performance comparison of federated learning methods on CIFAR10-LT, PathMNIST-LT, and OCTMNIST-LT across various test distributions.}
    \begin{tabular}{>{\centering\arraybackslash}p{0.10\textwidth}|%
                    >{\centering\arraybackslash}p{0.04\textwidth}%
                    >{\centering\arraybackslash}p{0.04\textwidth}%
                    >{\centering\arraybackslash}p{0.04\textwidth}%
                    >{\centering\arraybackslash}p{0.04\textwidth}%
                    >{\centering\arraybackslash}p{0.04\textwidth}|%
                    >{\centering\arraybackslash}p{0.04\textwidth}|%
                    >{\centering\arraybackslash}p{0.04\textwidth}%
                    >{\centering\arraybackslash}p{0.04\textwidth}%
                    >{\centering\arraybackslash}p{0.04\textwidth}%
                    >{\centering\arraybackslash}p{0.04\textwidth}%
                    >{\centering\arraybackslash}p{0.04\textwidth}|%
                    >{\centering\arraybackslash}p{0.05\textwidth}}
    \specialrule{1.2pt}{0pt}{0pt}
    \multicolumn{13}{c}{\textbf{CIFAR10-LT}} \\
    \toprule
    \multirow{2}{*}{\textbf{Method}} & \multicolumn{5}{c|}{\textbf{Forward-LT}} & \textbf{Uni.} & \multicolumn{5}{c|}{\textbf{Backward-LT}} & \multirow{2}{*}{\textbf{Avg.}} \\
    \cmidrule(lr){2-6} \cmidrule(lr){8-12}
    & 50 & 25 & 10 & 5 & 2 & 1 & 2 & 5 & 10 & 25 & 50 & \\
    \midrule
    FedAvg    & 42.15 & 37.57 & 31.61 & 26.55 & 20.20 & 15.75 & 11.77 & 7.48  & 5.17  & 2.97  & 1.93  & 18.47 \\
    FedProx   & 38.21 & 34.90 & 29.73 & 25.41 & 19.50 & 15.29 & 11.34 & 7.28  & 4.95  & 2.85  & 1.82  & 17.39 \\
    BalanceFL & 39.75 & 36.02 & 30.17 & 25.57 & 19.97 & 19.00 & 18.35 & 18.40 & 20.27 & 22.33 & 23.58 & 24.86 \\
    CReFF     & 79.64 & 78.26 & \textbf{75.24} & \textbf{73.61} & \textbf{71.99} & \textbf{71.60} & 71.52 & 72.05 & 72.38 & 71.91 & 72.34 & 73.69 \\
    FedGraB   & 73.20 & 69.06 & 62.56 & 56.43 & 47.77 & 41.07 & 34.80 & 27.30 & 22.18 & 16.23 & 12.88 & 42.13 \\
    FedLoGe   & 42.58 & 37.66 & 30.53 & 25.17 & 18.17 & 13.44 & 9.50  & 5.67  & 3.62  & 1.77  & 1.14  & 17.20 \\
    \midrule
    Ours & \textbf{80.61} & \textbf{78.94} & 75.20 & 73.47 & 71.19 & 70.80 & \textbf{72.09} & \textbf{74.08} & \textbf{75.10} & \textbf{75.57} & \textbf{78.14} & \textbf{75.02} \\
    \bottomrule

    \toprule
    \multicolumn{13}{c}{\textbf{PathMNIST-LT}} \\
    \toprule
    \multirow{2}{*}{\textbf{Method}} & \multicolumn{5}{c|}{\textbf{Forward-LT}} & \textbf{Uni.} & \multicolumn{5}{c|}{\textbf{Backward-LT}} & \multirow{2}{*}{\textbf{Avg.}} \\
    \cmidrule(lr){2-6} \cmidrule(lr){8-12}
    & 50 & 25 & 10 & 5 & 2 & 1 & 2 & 5 & 10 & 25 & 50 & \\
    \midrule
    FedAvg    & 50.35 & 46.67 & 41.60 & 36.94 & 31.45 & 26.45 & 25.69 & 24.58 & 23.28 & 21.17 & 19.72 & 31.63 \\
    FedProx   & 44.32 & 42.14 & 37.52 & 35.85 & 32.76 & 29.63 & 25.78 & 21.08 & 19.44 & 17.74 & 16.36 & 29.33 \\
    BalanceFL & 32.71 & 33.77 & 34.48 & 34.75 & 35.66 & 34.94 & 33.35 & 30.24 & 30.80 & 33.77 & 35.61 & 32.74 \\
    CReFF     & 78.19 & 77.62 & 76.56 & 75.63 & 74.44 & \textbf{74.57} & \textbf{74.54} & 75.63 & 76.24 & 76.81 & 77.26 & 76.13 \\
    FedGraB   & 21.69 & 22.48 & 24.24 & 24.89 & 25.69 & 25.40 & 24.88 & 22.89 & 20.80 & 17.94 & 16.36 & 22.48 \\
    FedLoGe   & 64.85 & 64.48 & 52.88 & 47.32 & 38.20 & 31.69 & 25.60 & 18.96 & 14.56 & 10.28 & 7.31  & 34.19 \\
    \midrule
    Ours & \textbf{83.87} & \textbf{81.75} & \textbf{79.76} & \textbf{76.85} & \textbf{76.21} & 72.96 & 73.99 & \textbf{76.02} & \textbf{76.32} & \textbf{77.92} & \textbf{79.12} & \textbf{77.71} \\
    \bottomrule

    \toprule
    \multicolumn{13}{c}{\textbf{OCTMNIST-LT}} \\
    \toprule
    \multirow{2}{*}{\textbf{Method}} & \multicolumn{5}{c|}{\textbf{Forward-LT}} & \textbf{Uni.} & \multicolumn{5}{c|}{\textbf{Backward-LT}} & \multirow{2}{*}{\textbf{Avg.}} \\
    \cmidrule(lr){2-6} \cmidrule(lr){8-12}
    & 50 & 25 & 10 & 5 & 2 & 1 & 2 & 5 & 10 & 25 & 50 & \\
    \midrule
    FedAvg    & 73.82 & 66.84 & 56.30 & 47.08 & 34.52 & 25.20 & 21.78 & 16.20 & 12.16 & 8.02  & 5.29  & 33.38 \\
    FedProx   & 75.29 & \textbf{69.25} & 60.81 & 51.22 & 38.77 & 29.50 & 21.51 & 16.01 & 11.94 & 7.75  & 5.29  & 35.21 \\
    BalanceFL & \textbf{75.29} & 68.45 & 58.78 & 50.28 & 37.53 & 28.50 & 35.21 & 45.95 & 54.95 & 64.97 & 71.47 & 53.76 \\
    CReFF     & 38.53 & 39.04 & 39.19 & 37.10 & 36.58 & 35.30 & 35.34 & 40.11 & 41.89 & 45.45 & 47.94 & 39.68 \\
    FedGraB   & 73.53 & 66.84 & 56.31 & 47.08 & 34.66 & 25.80 & 21.51 & 16.01 & 12.61 & 9.09  & 7.06  & 33.68 \\
    FedLoGe   & 73.53 & 66.84 & 56.31 & 47.08 & 34.25 & 25.00 & 17.12 & 9.42  & 5.63  & 2.67  & 1.47  & 30.85 \\
    \midrule
    Ours & 72.65 & 68.72 & \textbf{63.06} & \textbf{57.63} & \textbf{59.86} & \textbf{58.90} & \textbf{59.59} & \textbf{65.73} & \textbf{69.82} & \textbf{80.20} & \textbf{74.41} &\textbf{66.42 } \\
    \bottomrule
    \end{tabular}
    \label{main_result}
\end{table*}

\subsection{Main Results}

Table~\ref{main_result} presents comprehensive performance comparisons across three datasets under varying test distribution scenarios. The results validate our central hypothesis: GTAFL achieves robust test-agnostic performance by trading minimal accuracy in predictable scenarios for substantial gains under distribution shifts.

\subsubsection{Results on CIFAR10-LT}


Averaged across all 11 test distributions, GTAFL achieves 75.02\% accuracy, surpassing CReFF's 73.69\% accuracy and all other baselines by a clear margin, demonstrating consistent superiority regardless of test distribution type. 
When test distributions reverse training class frequencies, particularly in the Backward50 long-tail setting, GTAFL achieves 78.14\% accuracy, 
representing a remarkable 76\% improvement over FedAvg and a meaningful 6\% gain over CReFF. This substantial improvement validates our approach's 
effectiveness when deployment conditions fundamentally differ from training assumptions. In forward scenarios, GTAFL maintains competitive performance with CReFF while showing consistent improvements at moderate imbalance factors. 
At IF=50 and IF=25, GTAFL outperforms CReFF by 0.97\% and 0.68\% respectively, demonstrating stable performance across varying imbalance severities. Although GTAFL performs marginally below CReFF in uniform and some forward settings, 
the gap remains within 0.8\%, confirming competitive accuracy while maintaining 
substantially superior backward robustness.

\subsubsection{Medical Domain Validation}
GTAFL similarly demonstrates average performance advantages in medical imaging 
datasets, supporting its clinical applicability. As shown in Table~\ref{main_result}, GTAFL achieves 83.87\% at Forward50 on PathMNIST-LT, substantially outperforming all baselines, and attains nearly 2\% improvement over the best competing method in backward distributions. Averaged across all 11 test distributions on PathMNIST-LT, GTAFL achieves 77.71\%, confirming 
consistent superiority beyond individual distribution settings. On OCTMNIST-LT, while FedAvg and FedProx achieve marginally higher Forward50 accuracy, however, this advantage is misleading, as these methods simply assign test samples to the training majority classes, inflating forward accuracy while collapsing under backward distributions. By contrast, GTAFL maintains robust performance across the full distribution spectrum, achieving 74.12\% at Backward50, a 26.18\% 
improvement over FedAvg. These results demonstrate that GTAFL's test-agnostic design is particularly valuable for real-world federated healthcare applications where deployment distributions are unpredictable.

\subsection{\gy{Ablation Study}}

\begin{table*}[!ht]\small
    \centering
    \caption{\gy{Ablation study evaluating individual and combined effects of GTAFL 
    components on CIFAR10-LT. Components: CR = Classifier Retraining, GS = Global 
    Sampling, CL = Contrastive Learning, EN = Ensemble.}}
    \begin{tabular}{>{\centering\arraybackslash}p{0.09\textwidth}%
                    >{\centering\arraybackslash}p{0.04\textwidth}%
                    >{\centering\arraybackslash}p{0.04\textwidth}%
                    >{\centering\arraybackslash}p{0.04\textwidth}|%
                    >{\centering\arraybackslash}p{0.04\textwidth}%
                    >{\centering\arraybackslash}p{0.05\textwidth}%
                    >{\centering\arraybackslash}p{0.05\textwidth}|%
                    >{\centering\arraybackslash}p{0.05\textwidth}|%
                    >{\centering\arraybackslash}p{0.05\textwidth}%
                    >{\centering\arraybackslash}p{0.05\textwidth}%
                    >{\centering\arraybackslash}p{0.05\textwidth}|%
                    >{\centering\arraybackslash}p{0.05\textwidth}}
    \arrayrulecolor{black}
    \specialrule{1.2pt}{0pt}{0pt}
    \multicolumn{12}{c}{\gy{\textbf{CIFAR10-LT Ablation Study}}} \\
    \toprule
    \multicolumn{4}{c|}{\gy{\textbf{Components}}} & \multicolumn{3}{c|}{\gy{\textbf{Forward-LT}}} & \gy{\textbf{Uni.}} & \multicolumn{3}{c|}{\gy{\textbf{Backward-LT}}} & \multirow{2}{*}{\gy{\textbf{Avg.}}} \\
    \cmidrule(lr){1-4} \cmidrule(lr){5-7} \cmidrule(lr){9-11}
    \gy{\textbf{CR}} & \gy{\textbf{EN}} & \gy{\textbf{CL}} & \gy{\textbf{GS}} & \gy{\textbf{50}} & \gy{\textbf{25}} & \gy{\textbf{10}} & \gy{\textbf{1}} & \gy{\textbf{10}} & \gy{\textbf{25}} & \gy{\textbf{50}} & \\
    \midrule
    \gy{}           & \gy{}           & \gy{\checkmark} & \gy{\checkmark} & \gy{83.72} & \gy{79.28} & \gy{77.42} & \gy{66.32} & \gy{54.92} & \gy{47.20} & \gy{42.29} & \gy{64.45} \\
    \gy{\checkmark} & \gy{}           & \gy{}           & \gy{}           & \gy{79.64} & \gy{78.26} & \gy{75.24} & \gy{71.60} & \gy{72.38} & \gy{71.91} & \gy{72.34} & \gy{74.48} \\
    \gy{\checkmark} & \gy{}           & \gy{}           & \gy{\checkmark} & \gy{79.25} & \gy{77.98} & \gy{75.51} & \gy{70.56} & \gy{71.35} & \gy{71.14} & \gy{72.02} & \gy{73.97} \\
    \gy{\checkmark} & \gy{\checkmark} & \gy{}           & \gy{\checkmark} & \gy{79.64} & \gy{78.35} & \gy{75.76} & \gy{70.54} & \gy{71.23} & \gy{70.73} & \gy{71.45} & \gy{73.96} \\
    \gy{\checkmark} & \gy{\checkmark} & \gy{\checkmark} & \gy{}           & \gy{81.22} & \gy{79.03} & \gy{76.22} & \gy{71.51} & \gy{74.71} & \gy{76.18} & \gy{77.50} & \gy{76.62} \\
    \gy{\checkmark} & \gy{\checkmark} & \gy{\checkmark} & \gy{\checkmark} & \gy{80.61} & \gy{78.94} & \gy{75.20} & \gy{70.80} & \gy{75.10} & \gy{75.57} & \gy{78.14} & \gy{76.34} \\
    \bottomrule
    \end{tabular}
    \arrayrulecolor{black}
    \vspace{0.3cm}
    \label{ablation}
\end{table*}

\gy{We systematically analyze each component's contribution through controlled 
ablations on CIFAR10-LT across forward, uniform, and backward test scenarios. 
Table~\ref{ablation} reveals the distinct roles of our four components: 
Classifier Retraining (CR), Expert Ensemble (EN), Contrastive Learning (CL), 
and Global Sampling (GS)}.

\gy{Classifier Retraining (CR) establishes the fundamental capability for 
test-agnostic generalization. Without CR, the combination of CL and GS achieves 
competitive forward performance but suffers 41\% degradation under backward 
distributions. This confirms that self-supervised learning and global sampling 
alone cannot compensate for the distributional bias introduced by long-tailed 
training.}

\gy{The most significant improvement beyond CR emerges when Contrastive Learning 
(CL) is integrated with CR and EN: Backward50 performance increases from 
71.45\% to 78.14\%, a substantial 6.69\% gain. This validates that CL 
effectively acquires distribution-invariant representations, which are critical 
for generalizing to unknown test priors. Global Sampling (GS) provides 
measurable but more modest benefits. When combined with CR and EN, GS 
contributes consistent improvements on backward distributions such as 
Backward10 and Backward50.}

\gy{In summary, the ablation confirms that each component targets a distinct 
aspect of test-agnostic learning: CR addresses classifier bias through 
synthetic expert retraining, EN exploits complementary expert knowledge 
across distributional priors, CL strengthens feature robustness through 
self-supervised learning, and GS refines sampling balance to amplify the 
effectiveness of the other components. This systematic decomposition 
validates our architectural design for handling unknown test distributions 
in federated environments.}

\subsection{In-depth Analysis}
\subsubsection{Class-Wise Analysis}

\begin{table*}[t]\small
    \centering
    \caption{Top-1 test accuracy (\%) achieved by compared FL methods and CReFF on CIFAR10-LT Forward 50, Uniform, and Backward 50 settings.}
    \begin{tabular}{l|cccc|cccc|cccc}
    \toprule
    \textbf{CIFAR10-LT} & \multicolumn{4}{c|}{\textbf{Forward 50}} & \multicolumn{4}{c|}{\textbf{Uni.}} & \multicolumn{4}{c}{\textbf{Backward 50}} \\
    \midrule
    Method & All & Head & Middle & Tail & All & Head & Middle & Tail & All & Head & Middle & Tail \\
    \midrule
    FedAvg & 42.15 & 56.83 & 0.53 & 0.00 & 15.75 & 52.50 & 0.00 & 0.00 & 1.93 & 55.67 & 0.00 & 0.00 \\
    FedProx & 38.21 & 51.69 & 0.00 & 0.00 & 15.29 & 50.97 & 0.00 & 0.00 & 1.82 & 52.58 & 0.00 & 0.00 \\
    BalanceFL & 39.75 & 53.73 & 0.00 & 0.59 & 19.00 & 31.63 & 9.07 & 16.98 & 23.58 & 19.59 & 0.00 & 27.39 \\
    CReFF & 79.64 & 86.11 & 62.79 & \textbf{56.47} & \textbf{71.60} & 78.23 & \textbf{63.50} & 72.70 & 72.34 &\textbf{75.26} & 57.06 & 74.58 \\
    FedGraB & 73.20 & 87.08 & 40.79 & 11.18 & 41.07 & \textbf{78.57} & 40.33 & 13.50 & 12.88 & 72.16 & 39.34 & 6.33 \\
    FedLoGe & 42.58 & 57.55 & 0.18 & 0.00 & 13.44 & 44.73 & 0.07 & 0.00 & 0.14 & 32.99 & 0.00 & 0.00 \\
    \midrule
    Ours  & \textbf{80.61 } & \textbf{86.93}  & \textbf{65.65 }& 52.94  &70.8  &75.23 & 59.83 & \textbf{75.70}  & \textbf{78.14}  & 59.79  &\textbf{ 58.73}  & \textbf{81.90}  \\
    \bottomrule
    \end{tabular}
    
    \label{tab:class_wise}
\end{table*}

\gy{Table~\ref{tab:class_wise} presents a comprehensive class-wise performance breakdown on CIFAR10-LT, revealing how GTAFL adapts across different class categories under varying test priors. This granular analysis provides critical insights into the method's behavior when handling head, middle, and tail classes under distribution shifts.}

\gy{The most compelling finding emerges under backward conditions. GTAFL achieves 81.90\% tail-class accuracy in the backward setting, confirming that the model has successfully reallocated its representational capacity toward classes that are prevalent at test time. However, under test-agnostic condition, other models that maintain high accuracy on training-time head classes is failing to adapt to the new distribution. For example, CReFF produces higher head-class accuracy in backward50, this comes at the cost of overall performance, achieving only 72.34\% overall versus GTAFL’s 78.14\% and substantially lower tail-class accuracy. Furthermore, in clinical contexts such as epidemic outbreaks, identifying the newly dominant pathology is far more critical than maintaining precision on now-comparatively-rare conditions.}

\gy{Under forward test conditions, GTAFL achieves competitive or superior performance across all class categories. Notably, GTAFL surpasses CReFF on both head classes and middle classes, confirming that head-class reduction is not a systematic limitation. In uniform test scenarios, GTAFL delivers strong tail-class and head-class performance. Both performance validates GTAFL's ability to redistribute classification capacity effectively when test distribution is agnostic.}

\subsubsection{Robustness to Limited Training Data}
\label{data fraction}

\begin{figure}[!ht] 
    \centering
    \includegraphics[width=\linewidth]{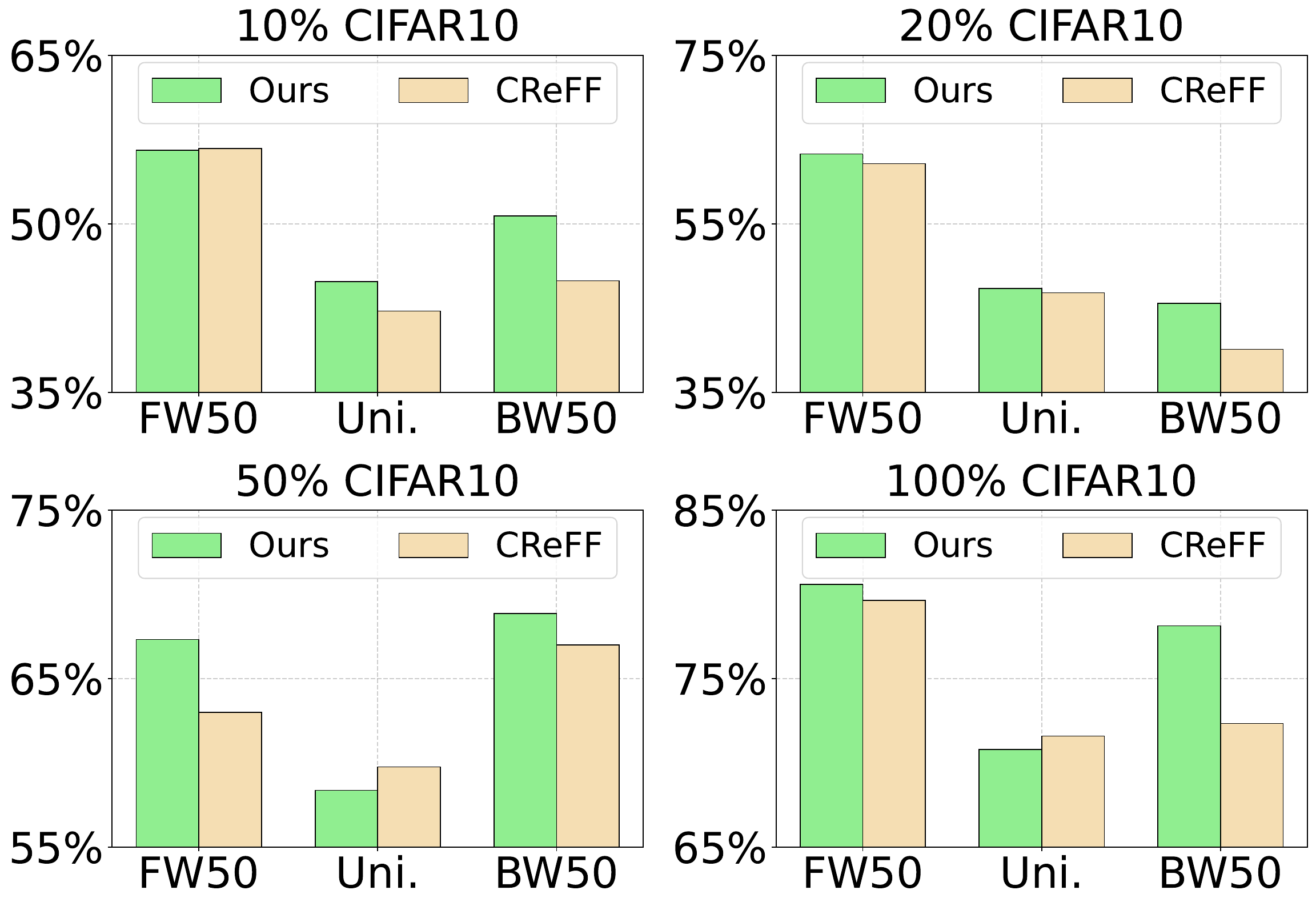} 
    \caption{Performance comparison of GTAFL and CReFF on varying CIFAR10 data fractions. GTAFL demonstrates greater robustness as the data size decreases. (FW50 = Forward50, Uni. = Uniform, BW50 = Backward50). When the data fraction reduces to 10\% of original dataset, GTAFL outperforms CReFF by nearly 5\%. For more details, please refer to \ref{data fraction}.}
    \label{fig:data_fraction} 
\end{figure}

To evaluate GTAFL's performance under data scarcity conditions prevalent in medical imaging scenarios, we conduct experiments with varying training data fractions (100\%, 50\%, 20\%, 10\%) on CIFAR10-LT. Figure~\ref{fig:data_fraction} presents the comparative analysis between GTAFL and CReFF across different data availability conditions.

The results show that  GTAFL consistently outperforms CReFF on backward distributions 
across all data fractions. For instance, at 10\% data, GTAFL achieves 50.70\% versus CReFF's 44.94\% on Backward50, demonstrating robust test-agnostic generalization even under severe data scarcity. This 
advantage stems from GTAFL's distribution-specialized expert classifiers, which maintain meaningful distributional diversity even when trained on limited federated features.

Furthermore, as training data decreases, GTAFL's advantage extends to 
uniform and forward distributions. At 50\% data, GTAFL surpasses CReFF on forward scenarios by 4.32\%, and at 10\% data, it also leads on uniform accuracy by 2.63\%. This suggests that GTAFL's server-side self-supervised refinement provides increasingly competitive feature representations as labeled data becomes more scarce. This is valuable for real-world medical federated learning, where annotation costs are high and labeled samples are often limited.

\subsubsection{\gy{Computational and Communication Overhead Analysis}}

\begin{table}[h]
\centering
\arrayrulecolor{black}
\caption{\gy{Computational overhead comparison between FedAvg, CReFF, and GTAFL on CIFAR10-LT. Client = Client Time, Server = Server Time, GFF Opt = GFF Optimization, CR = Classifier Retraining, CL = Contrastive Learning.}}
\label{tab:overhead_comparison}
\resizebox{\columnwidth}{!}{%
\begin{tabular}{lccccc}
\toprule
\gy{\textbf{Method}} & \gy{\textbf{Client}} & \gy{\textbf{Server}} & \gy{\textbf{GFF Opt}} & \gy{\textbf{CR}} & \gy{\textbf{CL}} \\
\midrule
\gy{FedAvg}       & \gy{1129s} & \gy{43s}   & \gy{—}    & \gy{—}     & \gy{—}     \\
\gy{CReFF}        & \gy{1376s} & \gy{597s}  & \gy{95s}  & \gy{502s}  & \gy{—}     \\
\gy{Ours}         & \gy{1712s} & \gy{4687s} & \gy{277s} & \gy{1586s} & \gy{2699s} \\
\bottomrule
\end{tabular}%
}
\arrayrulecolor{black}
\end{table}

\begin{table}[h]
\centering
\arrayrulecolor{black}
\caption{\gy{Communication overhead comparison over 200 training rounds on CIFAR10-LT.}}
\label{tab:comm_overhead}
\resizebox{\columnwidth}{!}{%
\begin{tabular}{lccc}
\toprule
\gy{\textbf{Method}} & \gy{\textbf{Uplink}} & \gy{\textbf{Downlink}} & \gy{\textbf{Extra vs.\ FedAvg}} \\
\midrule
\gy{FedAvg}       & \gy{7.333\,GB} & \gy{7.333\,GB} & \gy{—}          \\
\gy{CReFF}        & \gy{7.423\,GB} & \gy{7.349\,GB} & \gy{105\,MB}    \\
\gy{GTAFL (Ours)} & \gy{7.602\,GB} & \gy{7.380\,GB} & \gy{316\,MB}    \\
\bottomrule
\end{tabular}%
}
\arrayrulecolor{black}
\end{table}

\gy{To evaluate the practical feasibility of GTAFL, we compare its training and communication overhead against FedAvg and CReFF on CIFAR10-LT under the backward50 test distribution. As reported in Tables~\ref{tab:overhead_comparison} and~\ref{tab:comm_overhead}. GTAFL's client time represents a 24.4\% increase over CReFF, attributable solely to the additional backward passes required for computing 3 class-wise expert gradients. This is a lightweight operation making GTAFL practical for resource-constrained edge devices.}

\gy{The total server time is higher than CReFF , driven primarily by 
contrastive learning  refinement and extra classifier retraining. As shown in Table~\ref{tab:comm_overhead}, GTAFL requires only 316\,MB of additional communication compared 
to FedAvg over 200 communication rounds with 40\% participants each round . This marginal cost arises from uploading class-wise gradients for three expert classifiers and downloading the retrained expert heads, both of which involve only 
the lightweight classifier layer rather than the full model.}

\gy{In summary, GTAFL achieves its test-agnostic performance gains through a server-centric design that shields clients from additional complexity.}

\subsubsection{\gy{Ensemble Mechanism Analysis}}

\begin{table}[h]
\centering
\arrayrulecolor{black}
\caption{\gy{Comparison of ensemble strategies on CIFAR10-LT.
FW = Forward, BW = Backward, Conf.-Weighted = Confidence-Weighted.}}
\label{tab:ensemble_comparison}
\resizebox{\columnwidth}{!}{%
\begin{tabular}{lccccc}
\toprule
\gy{\textbf{Strategy}} & \gy{\textbf{FW50}} & \gy{\textbf{FW25}} & \gy{\textbf{Uni.}} & \gy{\textbf{BW25}} & \gy{\textbf{BW50}} \\
\midrule
\gy{Hard Voting (Ours)} & \gy{0.7456} & \gy{0.7321} & \gy{0.6749} & \gy{0.7684} & \gy{0.7660} \\
\gy{Soft Voting}        & \gy{0.7452} & \gy{0.7342} & \gy{0.6755} & \gy{0.7680} & \gy{0.7713} \\
\gy{Conf.-Weighted}     & \gy{0.7456} & \gy{0.7349} & \gy{0.6753} & \gy{0.7685} & \gy{0.7717} \\
\bottomrule
\end{tabular}%
}
\arrayrulecolor{black}
\end{table}

\gy{We further investigate the ensemble strategy by comparing our default hard 
voting, soft voting via averaged softmax probabilities, and confidence-weighted 
voting proportional to each expert's maximum predicted probability. As shown in 
Table~\ref{tab:ensemble_comparison}, all three strategies yield nearly identical 
performance across all test distributions, with maximum differences below 0.006. 
This negligible gap is theoretically expected: confidence-based weighting assumes 
that softmax probabilities are reliable expert-selection signals, but deep neural 
networks are known to produce poorly calibrated confidence scores under distribution 
shift~\cite{guo2017calibration}, causing experts to assign high confidence to 
incorrect predictions on unseen distributions. These results confirm that uniform 
hard voting is sufficient for test-agnostic federated learning.}

\subsubsection{Sensitivity Analysis of GFF Imbalance}

\begin{center}
    \includegraphics[width=1\columnwidth]{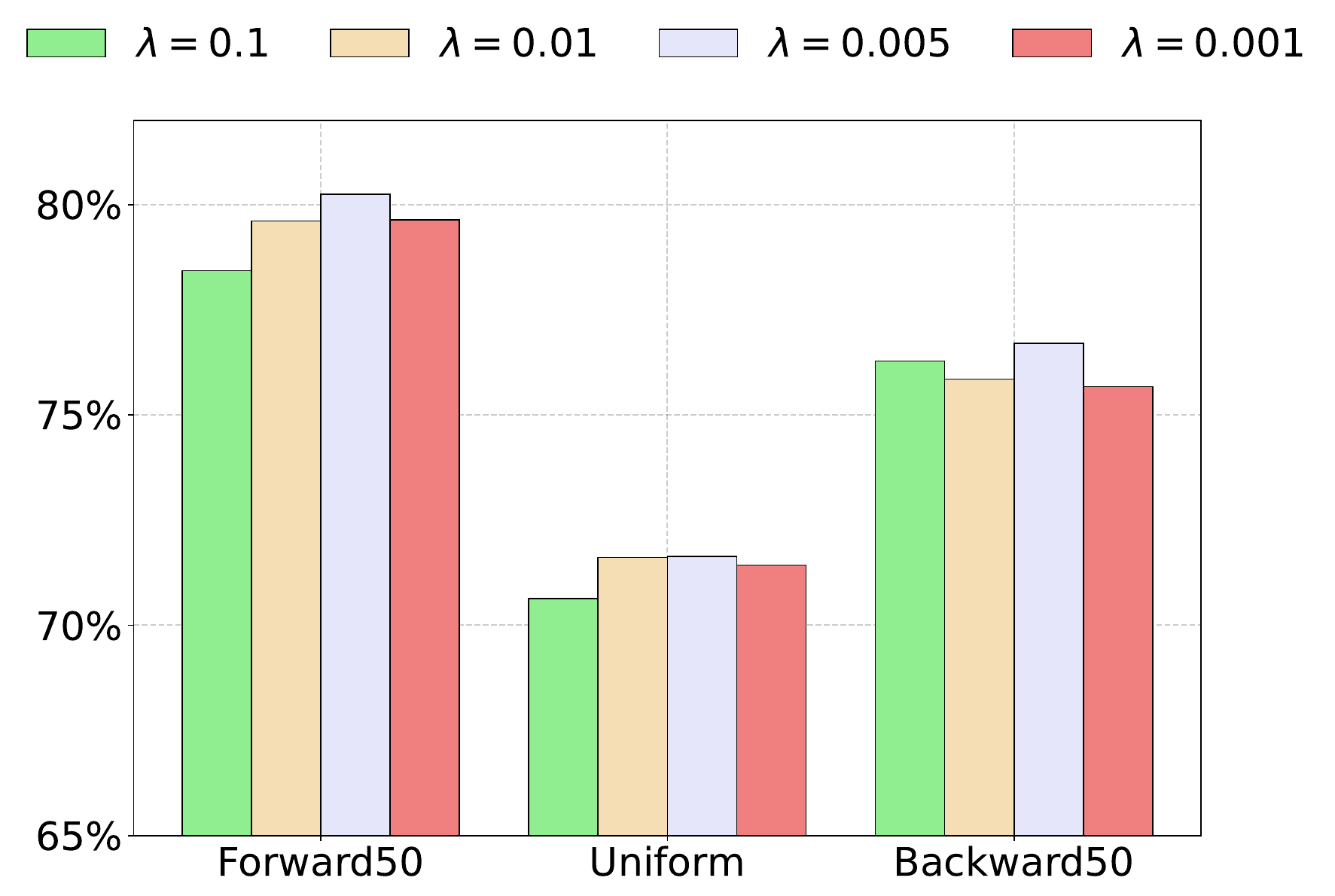}
    \captionof{figure}{\gy{Sensitivity analysis of GFF Imbalance on CIFAR10-LT. A larger $\lambda$ induces a more skewed GFF distribution across expert classifiers.}}
    \label{fig:lambda_sensitivity}
\end{center}

\gy{Figure~\ref{fig:lambda_sensitivity} reports the sensitivity of $\lambda$, and it controls the distributional imbalance of the synthetic 
global federated features (GFF) used for expert classifier retraining. When $\lambda$ is too small, all three experts degenerate into equivalent uniform classifiers, losing distribution-specific specialization. When $\lambda$ is too large, such as $\lambda = 0.1$, overly skewed synthetic features 
introduce distributional mismatch, degrading forward and uniform performance. The value $\lambda = 0.005$ achieves the best overall balance. Across $\lambda \in [0.001, 0.01]$, performance variations 
remain below 2\%, confirming that GTAFL is robust to moderate perturbations of this parameter.}

\section{Related Work}    
\subsection{Long-tail distribution FL}
Long-tail learning has recently attracted significant attention due to the prevalence of imbalanced data in real-world applications. In the context of federated learning, the long-tail problem becomes even more pronounced as it is exacerbated by data heterogeneity across local clients. Several solutions, such as re-sampling, representation learning, and classifier re-training, which are commonly employed in centralized learning, can also be adapted in federated setting. Re-sampling strategies assign different training weights to classes, over-sample minority classes, and under-sample majority classes based on their local frequencies to achieve a balanced training process \cite{shuai2022balancefl, cao2019learning}. Recent methods often decouple the training process into two stages:  classifier re-training and representation learning. During the classifier re-training phase, techniques such as post-calibration \cite{Kang2019DecouplingRA, wang2020devil, li2020overcoming} are applied to address class imbalance. In the federated learning scenario, CReFF\cite{shang2022federated} introduces federated features, a form of synthetic data, to re-train biased classifiers. Some studies have observed that balancing the feature space is critical for addressing long-tail problems. For instance, RUCR \cite{huang2024federated} employs global unified prototypes to represent class-wise characteristics and adjust the distorted feature space. Meanwhile, Moon \cite{Li_2021_CVPR} leverages contrastive learning to align the current local model with the global model while distancing it from previous local models. However, most existing methods are designed under the assumption of a uniform test distribution, which may not hold in real medical scenarios. 

\subsection{Test-agnostic learning}
Test-agnostic long-tailed tasks have gained significant attention in centralized training. Existing methods address varying test distributions by adjusting loss functions or training multiple experts tailored to different scenarios \cite{guo2024prototype, Aimar2022BalancedPO}. For example, LADE \cite{Hong2020DisentanglingLD} uses prior knowledge of test distributions for post-adjustment, though this is impractical in real-world settings. SADE \cite{Zhang2021SelfSupervisedAO} trains experts with specialized loss functions and ensembles them using unsupervised learning strategies. While effective in centralized settings, these approaches face challenges in federated learning. Local clients often lack the computational resources to train multiple models, and data heterogeneity can result in missing classes, leading to degraded global performance. Building on the idea of training and ensembling experts, GTAFL retrains expert classifiers on the server and ensembles them during inference, avoiding the need for resource-intensive training on local clients. Local models are trained following standard FedAvg \cite{mcmahan2017communication}, overcoming the limitations of centralized strategies in federated environments. While prior test -agnostic federated learning work has focused on challenges like fairness\cite{Yang2024HarnessingHL}, to the best of our knowledge, we are the first to identify and formulate the test-agnostic long-tailed federated learning, which presents fundamentally different challenges from existing problem formulations.

\section{Discussion}

\subsection{Privacy and Security Considerations}
GTAFL operates within the standard federated learning paradigm, requiring exchanges of model parameters and gradients between clients and server. While raw data remains localized, such updates can leak information under sophisticated adversaries. Previous work has demonstrated that gradient and model inversion attacks may partially reconstruct sensitive training data \cite{geiping2020inverting,yin2021see,kaissis2021end,weizhengFL}, which is particularly concerning in healthcare applications.
Our framework is compatible with privacy-preserving mechanisms to mitigate these risks. For example, client-side differential privacy (DP) \cite{wei2020federated,10589503,weizhengDP} can be applied to expert gradients and model updates via per-example clipping and calibrated noise, providing formal privacy guarantees with controllable utility-privacy trade-offs. 

\subsection{\gy{Performance Analysis and Trade-offs}}
\gy{GTAFL delivers substantial gains when test distribution differs from training setting, while maintaining competitive accuracy in other settings. This aligns with our design objective of test-agnostic generalization. However, a current limitation is relatively modest accuracy on majority classes under backward distribution conditions, reflecting an inherent trade-off when optimizing for substantial shifts in class priors. In clinical practice, this trade-off may be acceptable when deployment scenarios emphasize previously rare pathologies, such as emerging diseases or underdiagnosed conditions gaining recognition. Nevertheless, improving head-class fidelity while preserving tail-class robustness remains an important optimization target for comprehensive clinical deployment.}

\subsection{Future Directions and Clinical Translation}
Future work will focus on three directions. First, privacy-by-design instantiations: integrating DP with adaptive noise schedules for expert gradients, combined with secure aggregation, to obtain end-to-end privacy guarantees tailored to class-wise expert updates. \gy{Second, we aim to improve the efficiency of contrastive learning refinement by incorporating more computationally efficient contrastive objectives\cite{tan2022federated,yu2023multimodal}, reducing training overhead while preserving the representation quality needed for robust feature learning. Third, we find that excessively large $\lambda$ values can cause GFF gradient matching to degrade due to extremely severe mismatch, leading to reversed expert specialization. Future work will investigate more robust GFF optimization strategies and adaptive ensemble mechanisms that can automatically detect and down-weight degenerated expert classifiers at test time, improving reliability under suboptimal hyperparameter configurations. We plan to evaluate GTAFL under broader medical datasets and cross-site acquisition variability, and report privacy-utility trade-offs across different DP budgets to support clinical adoption.}

\section{Conclusion}
We introduce Test-Agnostic Long-Tailed Federated Learning and propose GTAFL to address distribution shifts through expert classifiers ensemble. GTAFL generates diverse global federated features for privacy-preserving expert retraining, demonstrating robust performance across distribution shifts without serious degradation on reverse or imbalanced distributions. Extensive experiments show GTAFL outperforms state-of-the-art long-tailed federated learning methods, validating the framework's effectiveness for healthcare applications where both privacy preservation and robustness to distribution shifts are critical.


\section*{References}
\bibliographystyle{IEEEtran}
\bibliography{IEEEabrv,Reference}

\end{document}